\pdfoutput=1

\documentclass[11pt]{article}

\usepackage[final]{acl}

\usepackage{times}
\usepackage{latexsym}

\usepackage[T1]{fontenc}

\usepackage[utf8]{inputenc}

\usepackage{microtype}

\usepackage{inconsolata}

\usepackage{graphicx}
\usepackage{amsmath}
\usepackage{amssymb}
\usepackage{bbm}
\usepackage{booktabs}
\usepackage{multirow}

\title{Reasoning Like a Doctor: Improving Medical Dialogue Systems via Diagnostic Reasoning Process Alignment}

\author{Kaishuai Xu$^{1}$, Yi Cheng$^{1\ast}$, Wenjun Hou$^{1,2\ast}$, Qiaoyu Tan$^{3}$, Wenjie Li$^{1}$ \\
$^1$The Hong Kong Polytechnic University, HKSAR, China \\
$^2$Southern University of Science and Technology, Shenzhen, China \\
$^3$New York University Shanghai, Shanghai, China\\
\texttt{\{kaishuaii.xu, alyssa.cheng\}@connect.polyu.hk, houwenjun060@gmail.com} \\
\texttt{qiaoyu.tan@nyu.edu, cswjli@comp.polyu.edu.hk}
}

\begin{document}
\maketitle
\begingroup\def\thefootnote{$\ast$}\footnotetext{Equal Contribution.}\endgroup
\begin{abstract}

Medical dialogue systems have attracted significant attention for their potential to act as medical assistants. Enabling these medical systems to emulate clinicians’ diagnostic reasoning process has been the long-standing research focus. Previous studies rudimentarily realized the simulation of clinicians’ diagnostic process by fine-tuning language models on high-quality dialogue datasets. Nonetheless, they overly focus on the outcomes of the clinician's reasoning process while ignoring their internal thought processes and alignment with clinician preferences. 
Our work aims to build a medical dialogue system that aligns with clinicians' diagnostic reasoning processes. We propose a novel framework, \textsc{Emulation}, designed to generate an appropriate response that relies on abductive and deductive diagnostic reasoning analyses and aligns with clinician preferences through thought process modeling.
Experimental results on two datasets confirm the efficacy of \textsc{Emulation}. Crucially, our framework furnishes clear explanations for the generated responses, enhancing its transparency in medical consultations.\footnote{Our codes are available at \url{https://github.com/kaishxu/Emulation}.}

\end{abstract}
\section{Introduction}

\begin{figure}[t!]
	\centering
	\includegraphics[width=0.75\linewidth]{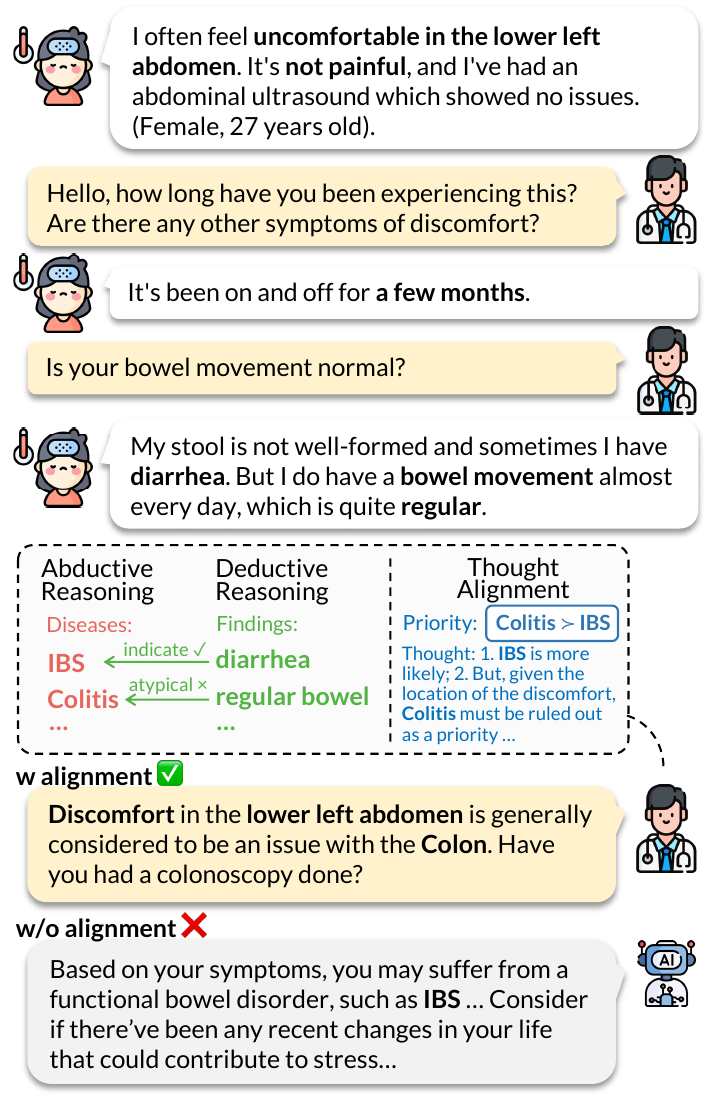}
	\caption{An example of a medical dialogue. \textbf{IBS} is the abbreviation for irritable bowel syndrome.}
	\label{intro_example}
\end{figure}

Medical dialogue systems as a fundamental tool in facilitating effective healthcare consultations have garnered sustained attention in recent years \cite{meddialog, meddg}. Especially with the rise of large language models (LLMs), these systems have shown promising potential to act as virtual medical assistants that can aid clinicians in accurate and efficient diagnosis \cite{disc_medllm, huatuo2, gpt-4}. 
In this research area, how to enable these medical systems to emulate clinicians’ diagnostic reasoning process has been the long-standing research focus ever since its inception \cite{holyoak2005cambridge}. 
To achieve this, previous research has paid tremendous efforts in constructing high-quality datasets of medical consultation dialogues, based on which they fine-tuned language models, rudimentarily realizing the simulation of clinicians’ diagnostic process that can inquire about symptoms and make a diagnosis \cite{disc_medllm, huatuo2, meddg, dfmed}. 

Nonetheless, existing research remains focused on the ``output'' of the clinician’s reasoning process (e.g., the symptoms and diseases mentioned in clinicians’ utterances), still neglecting the clinician's internal thought process and decision-making mechanisms. As a result, available systems overly rely on the co-occurrence patterns in the training data, prone to inquire about the most frequent symptoms and diagnose the most common disease. 
In fact, a real clinician’s thought process goes far beyond this. In clinical medicine, there is a concept known as ``Clinician Preference'' \cite{skills_for_medical_com,holyoak2005cambridge,clinical_preference}. It refers to the inclinations that healthcare professionals exhibit during diagnostic reasoning. It has been widely acknowledged that clinician preferences are affected by many factors \cite{skills_for_medical_com}, much more than the symptom and disease frequency. For example, in Figure \ref{intro_example}, the clinician prefers to discuss examinations in the next response to rule out colitis rather than consider factors that cause irritable bowel syndrome. This is because the location of the discomfort typically indicates a more severe issue in the colon. For the existing systems that are constructed purely through dialogue data fine-tuning, it is difficult to capture such subtle reasoning preferences \cite{doctorglm}. 

In this paper, we aim to develop a medical dialogue system that can align with the internal diagnostic reasoning process of clinicians. To this end, we must first model a diagnostic analysis process. For the multi-turn medical dialogue, clinicians generally adopt an iterative, abductive, and deductive analysis that discovers an explanation of the patient's condition and evaluates the effectiveness of the explanation \cite{holyoak2005cambridge}. This analysis provides a robust and comprehensive foundation for accurate diagnoses. Then, we need to align response generation with the clinician preference based on the analysis and dialogue context. The thought process of how clinicians reason and generate responses is a vital resource for learning their preferences. Extracting the thought process and modeling it together with responses can help learn the diagnostic reasoning process. 

Based on the above motivation, we propose a novel medical dialogue system framework, \textsc{Emulation}, which emulates clinicians' diagnostic reasoning processes to generate an appropriate response that relies on ample diagnostic analysis and aligns with clinician preferences in consultation. First, an abductive reasoning module investigates potential diseases that can explain a patient's condition. Then, a deductive reasoning module comprehensively analyzes the relation between clinical findings and potential diseases. Finally, the thought alignment module adjusts the potential disease priority that may be discussed next and generates thought processes that align with the clinician preference based on the above analysis. To learn the clinician preference, we build a new diagnostic thought process dataset with the help of an LLM. 

Our key contributions are outlined as follows: 
(1) We propose a novel medical dialogue system framework, \textsc{Emulation}, that emulates clinicians' diagnostic reasoning processes and aligns with clinician preferences. This is the first work that explores clinician preferences and internal thought processes.
(2) We build a diagnostic thought process dataset that is employed to align response generation with clinician preferences.
(3) Experimental results demonstrate the effectiveness and explainability of \textsc{Emulation}.

\section{Preliminary}

\subsection{Problem Formulation}

In our work, we conceptualize a medical dialogue as a sequence $U=\{(U^P_k, U^D_k)\}_{k=1}^T$, where each pair $(U^P_k, U^D_k)$ comprises an utterance from a patient followed by an utterance from a doctor. Each doctor's utterance is annotated with a list of diseases $E_t=\{e_{i}\}$ that could be relevant to the patient's condition mentioned in the dialogue. Given a dialogue history $U_t=\{U^P_1, U^D_1, ..., U^P_t\}$ up to the $t$-th patient utterance, our system's objective is to generate a contextually appropriate and medically informed $t$-th doctor's utterance $U^D_t$. 

\subsection{Diagnostic Reasoning}

Diagnostic reasoning involves a detailed examination of how doctors think and make decisions in medicine \cite{holyoak2005cambridge}. It serves as a foundational element for advanced cognitive activities, such as formulating diagnostic conclusions and grasping the underlying pathology of diseases. The framework of diagnostic reasoning has been a focal point of medical cognition research \cite{patel1997cognitive}. A widely recognized perspective holds that diagnosis represents an iterative, abductive, and deductive process of formulating and evaluating potential explanations for a patient's abnormal condition \cite{elstein1978medical, holyoak2005cambridge}. We summarize this perspective into two reasoning processes: \textbf{Abductive Reasoning} aims to create a plausible diagnosis to explain observed clinical findings; \textbf{Deductive Reasoning} further tests the available diagnoses by determining whether the findings support, refute, or are unrelated to the diagnoses. In our method, the first process efficiently explores a disease knowledge base to identify several possible diseases that explain the patient's condition in the current turn of the conversation. The second process comprehensively inspects the relationship between the clinical findings and possible diseases. These two processes are conducted iteratively across the conversation. Besides, the clinician preference for medical decision analysis plays an essential role in the diagnostic reasoning process \cite{skills_for_medical_com}. This preference highlights behaviors and actions in medical conversations unique to each clinician, which LLMs may lack. It is the nuanced preference difference that distinguishes a superior clinician from other clinicians \cite{medical_value}. We model a \textbf{Thought Alignment} process to align response generation with the general clinician preference in consultation.

\subsection{Disease Annotation}

Given GPT-4's demonstrated effectiveness in several medical licensing examinations \cite{gpt-4}, we employ it to annotate potential diseases relevant to the patient's condition automatically. We implement two ways to generate lists of diseases considered \textbf{before} and \textbf{after} examining the current doctor's response. For the first list, we construct a prompt with the dialogue history $U_t$, utilizing the model's diagnostic skills to identify potential diseases. For the second list, we create a prompt that incorporates both the history $U_t$ and the ground truth response $U^D_t$, enabling the model to deduce the diseases that doctors might discuss. Then, we link the inferred diseases with those in an external medical knowledge base and obtain two lists of potential diseases, $E^{pri}_t$ and $E^{post}_t$. We merge two lists as one $E_t$ for each doctor’s response. The details are described in the Appendix \ref{disease_annotate}.

\section{Method}

\begin{figure*}[t!]
	\centering
	\includegraphics[width=0.95\linewidth]{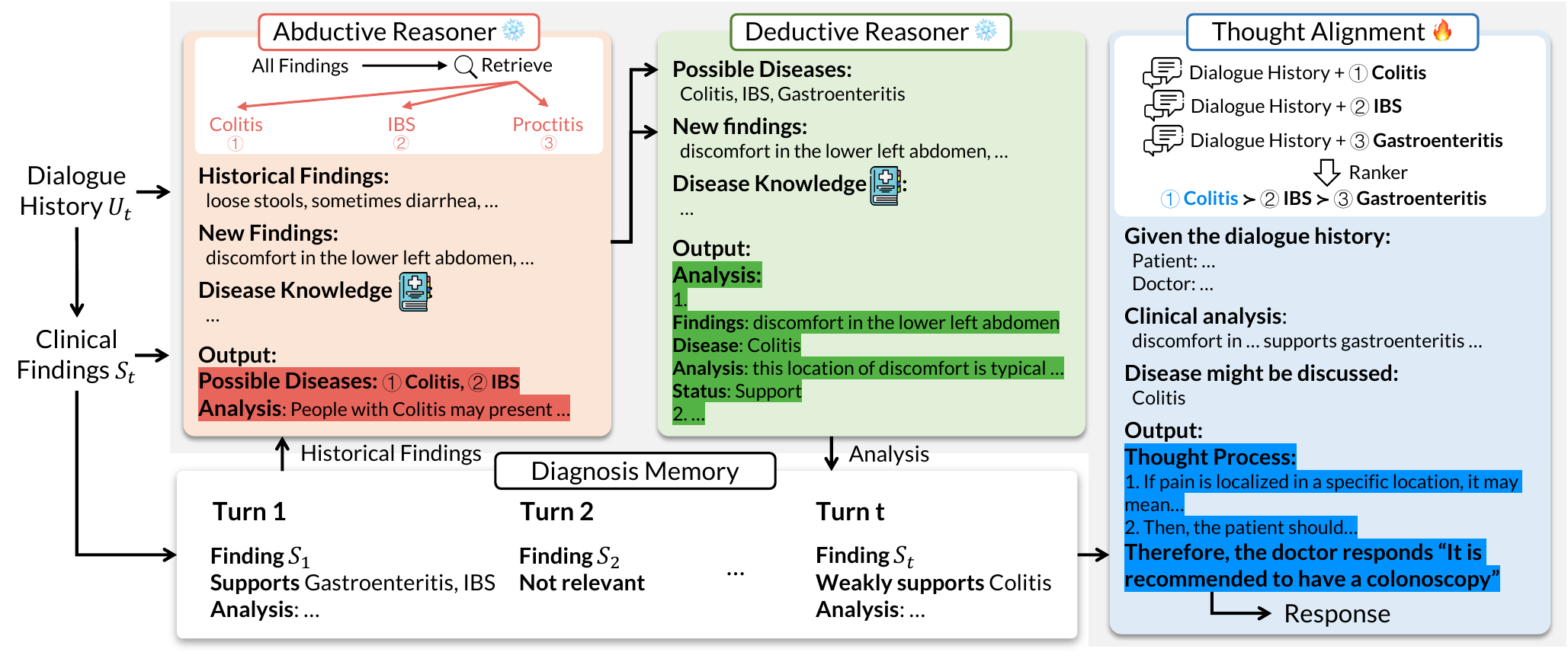}
	\caption{Illustration of our Framework \textsc{Emulation}. The symbols of a snowflake or a flame represent the module mainly operating in a prompt-based way or having undergone fine-tuning, respectively.}
	\label{framework}
\end{figure*}

In this section, we introduce a diagnostic reasoning framework, which analyzes patient conditions through abductive reasoning and deductive reasoning \cite{holyoak2005cambridge}, and aligns responses with the clinician preferences. As shown in Figure \ref{framework}, our framework includes three modules. The \textit{Abductive Reasoner} (\S \ref{abductive}) first generates potential diseases based on the new clinical findings of each turn. Then, the \textit{Deductive Reasoner} (\S \ref{deductive}) comprehensively analyzes the relation between potential diseases and new clinical findings. Finally, the \textit{Thought Alignment} (\S \ref{align}) module adjusts the disease priority that might be discussed in subsequent dialogues and performs thought process modeling to generate the next response. 

Given the medical dialogue history $U_t$ at the $t$-th conversation turn, we first use a pre-trained LLM to summarize new clinical findings from the recent utterances $U^D_{t-1}$ and $U^P_t$. The clinical findings are presented using the SOAP note \cite{soap}, a documentation technique utilized by healthcare professionals for recording notes (e.g., chief complaint) in a patient's medical record. For example, we provide the LLM with a prompt to ``\textit{summarize the clinical findings from the recent conversation between doctor and patient, adhering to the structure of the SOAP note}''. Each of the summarized clinical findings is in the phrase format and merged into a set $S_t = \{s_{j}\}_{j=1}^{m}$, where $m$ represents the number of clinical findings at the $t$-th conversation turn.

\subsection{Abductive Reasoner}\label{abductive}

The Abductive Reasoner aims to find the most possible diseases that explain the patient's abnormal condition. Previous studies rely solely on the diagnosis capability of black-box language models or focus on limited diseases in one dataset. We design a two-step pipeline to achieve a comprehensive and explainable abduction. The first step employs a disease retriever to narrow down the scope of the disease. Then, the second step leverages a pre-trained LLM to generate a potential disease list with detailed explanations based on the clinical findings and external medical knowledge.

\paragraph{Disease Retrieval.}
We concatenate all findings as the query and retrieve disease documents that are relevant to the findings from an external medical knowledge base. A dense retriever with a BERT encoder is applied to collect relevant documents. The sequences of the query and disease document, each prefixed with a ``[CLS]'' token, are fed independently into the encoder. The hidden state corresponding to the ``[CLS]'' token from each sequence is chosen to serve as their respective representations $\mathbf{h} \in \mathbb{R}^d$ and $\mathbf{h}_{doc} \in \mathbb{R}^d$. We compute relevance scores through dot product as $\mathbf{h}^\mathsf{T} \cdot \mathbf{h}_{doc}$. The diseases corresponding to the top-$K$ documents ranked by the scores are considered candidate diseases ${\hat{E}_t}^0 = \{e_{i}\}_{i=1}^K$.
Our approach utilizes Contrastive Learning \cite{simcse} to train the retrieval model. The loss function applied to the training is specified as follows:
\begin{flalign}
&{\cal{L}}_{\text{CL}} = \nonumber & \\ 
& - \log \frac {\exp \left(  \mathbf{h}^\mathsf{T} \! \cdot \! \mathbf{h}_{doc}^+ \right) }{ \exp \left(  \mathbf{h}^\mathsf{T} \! \cdot \! \mathbf{h}_{doc}^+ \right) \! + \! \mathop{\sum}\limits_{\mathcal{C}} \exp \left( \mathbf{h}^\mathsf{T} \! \cdot \! \mathbf{h}_{doc}^- \right) } , &
\end{flalign}
where $\mathbf{h}_{doc}^+$ and $\mathbf{h}_{doc}^-$ denotes document representations of the relevant diseases in this turn $E_t$ and the irrelevant ones in the knowledge base $\mathcal{C}$. We adopt in-batch negative sampling for training. 

\paragraph{Diagnosis Refinement.}
After the retrieval, we have assembled a list of potential diseases; however, this approach unavoidably results in the inclusion of diseases irrelevant to the patient's condition. Furthermore, the disease retrieval, based on vector similarity, can not elucidate the rationale behind the presence of specific diseases within the potential list. In pursuit of an explainable diagnosis, we utilize a pre-trained LLM to identify possible diseases from this list, offering an explanation grounded in clinical findings. The diagnosis knowledge of each disease is provided since it can appropriately reduce the hallucination issue of LLMs. For instance, we prompt the LLM with ``\textit{given the findings and knowledge, select possible diseases that can explain the new clinical findings while satisfying past findings}''. To achieve a stable disease identification, we adopt a majority vote inspired by \citet{consist_cot}. Specifically, we randomly divide the list into batches of the same size (since the maximum input tokens of LLMs limit the diseases and knowledge that can be input). Such division is repeated several times to construct different batch groups. We utilize batch groups to generate several refined disease lists and calculate the voting score of each disease as follows:
\begin{equation}
v(e_i) = \sum_{j=1}^B \mathbbm{1}(e_i \in {\hat{E}_t}^j), e_i \in {\hat{E}_t}^0,
\end{equation}
where $B$ stands for the number of batch groups, and ${\hat{E}_t}^j$ represents the $j$-th refined list. The final refined list includes diseases with a voting score exceeding $B/2$, denoted as ${\hat{E}_t}' = \{ e_i \}_{i=1}^{K'}$.

\subsection{Deductive Reasoner}\label{deductive}

While the Abductive Reasoner has offered an analysis of the relationship between potential diseases and new clinical findings, this analysis predominantly aims to confirm a disease rather than exclude one or sift through information irrelevant to the current diagnosis. It is crucial for clinicians to thoroughly examine the information presented in the dialogue history. However, previous studies often overlook these non-affirmative analyses in their diagnostic processes, which is not conducive to improving diagnostic accuracy. 

We introduce the Deductive Reasoner to evaluate the affirmative and non-affirmative relations between new clinical findings and possible diseases. Given the clinical findings $S_t$ and the refined disease list ${\hat{E}_t}'$, this reasoner applies a pre-trained LLM with a prompt as ``\textit{analyze if the new clinical findings support, oppose, or are irrelevant to the possible diseases}''. We improve the accuracy of analysis by incorporating additional diagnostic knowledge of each disease. The content of specific tags, i.e., findings, disease, and status, is extracted as diagnosis memory $M_t$ for subsequent response generation.

\subsection{Thought Alignment}\label{align}

The clinician preference in consultation is critical, as it differentiates an experienced clinician from a novice clinical student. Prior research has always concentrated on diagnosing the most likely disease and discussing extra symptoms or treatment based on that diagnosis, which does not accurately reflect the dynamics of an actual healthcare consultation. To address this, we have developed the Thought Alignment module, which is designed to adapt preferences akin to that of an expert clinician. The module first prioritizes diseases to be discussed in the subsequent dialogue. Then, it models the thought process and response of real clinicians to align with clinician preferences.

\paragraph{Disease Priority Alignment}
We implement a disease ranker to establish the priority of each disease within the refined disease list ${\hat{E}_t}'$. The backbone of this ranker is the BERT encoder. For the input, we combine the dialogue history with each potential disease, formatting it as follows: ``[CLS] \{history\} \textit{the next response will discuss:} \{disease\}''. The relevance score of the dialogue history to the discussed disease is calculated:
\begin{equation}
r(U_t, e_i) = \mathtt{MLP}(\mathtt{repr}(U_t ; e_i)), e_i \in {\hat{E}_t}',
\end{equation}
where $\mathtt{repr}(\cdot)$ extracts the hidden state of the ``[CLS]'' token as the representation, and $\mathtt{MLP}(\cdot)$ projects the representation to a scalar. The ranking model is also trained using a contrastive loss function. To better adjust the priority within the refined disease list, the selection of negative samples includes the union of this refined list ${\hat{E}_t}'$ and the pre-annotated disease list $E_t^{pri}$. The positive diseases are from $E_t^{post}$. We compute relevance scores between the dialogue history and all diseases in the refined list and then reorder the list to ${\hat{E}_t}''$. 

\paragraph{Thought Process Alignment}
Thought processes are a reflection of clinician preferences in consultation. Motivated by studies on distilling multi-step reasoning capabilities \cite{slm_reason, dialog_cot}, we design a thought distillation method to empower a model that can generate thought processes consistent with clinician preferences. Specifically, we first leverage a pre-trained LLM to deduce a plausible thought process for each doctor's response. The prompt to the LLM is organized as ``\textit{complete the thought process based on the dialogue context}''. We augment the prompt with 3 human-annotated thought processes in the Chain-of-Thought (CoT) format. The thought extraction can be defined as:
\begin{equation}
Y_t \sim \mathtt{LLM}(Y_t|U_t, U_t^D, \text{prompt}).
\end{equation}
Then, we gather the thought process for each dialogue turn and develop a thought process alignment model using the autoregressive language modeling approach as our training objective. The loss function can be defined as:
\begin{equation}
\mathcal{L}_{\text{G}} = \! - \! \sum_j \log p(y_{jt}|U_t, M_{\leq t}, E^{post}_t, Y_{t, <j}),
\end{equation}
where $M_{\leq t}$ denotes the diagnosis memory until the $t$-th turn and $E^{post}_t$ denotes the diseases discussed in the next response. During the inference, we utilize the top-$K''$ diseases in the list ${\hat{E}_t}''$ instead of $E^{post}_t$ and their related analyses to generate the thought process. The final response is extracted from ``\textit{Therefore, the doctor responds} \{response\}''.

\section{Thought Process Corpus}

\begin{table}[t!]
\center
\resizebox{0.75\linewidth}{!}{
\begin{tabular}{@{}lll@{}}
\toprule
\textbf{Description}             & \multicolumn{1}{c}{\textbf{MedDG}} & \multicolumn{1}{c}{\textbf{KaMed}} \\ \midrule
\# Thought processes    & \multicolumn{1}{c}{122K}  & \multicolumn{1}{c}{285K}  \\
Avg. \# steps           & \multicolumn{1}{c}{4.17}  & \multicolumn{1}{c}{4.13}  \\
Avg. \# tokens per step & \multicolumn{1}{c}{40.98} & \multicolumn{1}{c}{41.21} \\
Avg. \# total tokens    & \multicolumn{1}{c}{236.39}& \multicolumn{1}{c}{239.53} \\ \bottomrule
\end{tabular}
}
\caption{Statistics of MEDIATOR}
\label{thought_corpus}
\end{table}

\begin{table}[!t]
\centering
\small
\begin{tabular}{p{0.9\linewidth}}
\toprule
\textbf{Patient}: What's going on with my throat feeling itchy and making me want to gag? (Female, 22 years old) \\
\textbf{Doctor}: How long have you been experiencing this?  \\
\textbf{Patient}: About half an hour. And my temperature is just under 37 degrees. \\
\textbf{Doctor}: Have you had pharyngitis before?  \\
\textbf{Patient}: No, but I have chronic gastritis. \\ 
\textbf{Doctor}: So, this hasn't happened before? \\
\textbf{Patient}: \underline{Only when} I have a \underline{cold}, I would feel \underline{itchy} and \underline{cough}. Also, I tend to feel uncomfortable when the \underline{seasons change}. Other than that, no. \\
\midrule
\textbf{Thought Processs}: \\
1. The patient mentions symptoms of itchy throat and gagging during \underline{colds} or when the \underline{seasons change}. This indicates respiratory issues. \\
2. The doctor considers the possibility of an \underline{allergic} reaction, as ... \underline{Seasonal changes} frequently introduce variations in \underline{allergens and irritants} in the environment, which might explain the patient's symptoms. \\
3. The patient's symptoms are \underline{not frequent}, but appear to be triggered by specific conditions or events. This further supports the allergies or environmental causes.  \\
4. The patient mentions having chronic gastritis, but the relation to the current symptoms is unclear ... but usually does not cause an itchy throat. So, confirming \underline{allergy issues} is a priority. \\
\textbf{Therefore, the doctor responds}, "Have you done any cleaning today? Or been exposed to dust mites?"... \\
\bottomrule
\end{tabular}
\caption{An example of the generated thought process.}
\label{thought_example}
\end{table}

\begin{table}[t!]
\center
\resizebox{0.88\linewidth}{!}{
\begin{tabular}{cccc}
\toprule
\textbf{Datasets} & \textbf{Knowledge}  & \textbf{Consistency} & \textbf{Rationality} \\ \midrule
MedDG    & 90\%        & 89\%         & 82\%         \\
KaMed    & 87\%        & 91\%         & 79\%         \\ \bottomrule
\end{tabular}
}
\caption{Human evaluation results of automatically generated thought processes. Values in the table represent average valid percentages.}
\label{human_cot}
\end{table}

In this section, we present the medical dialogue thought process corpus, MEDIATOR, where each dialogue turn is annotated with a chain-of-thought reasoning path. We utilize the powerful reasoning ability of GPT-4 and its impressive professional skills in the medical domain for automatic annotation. The annotated reasoning path is expected to reflect the thought process of clinicians who analyze the patient's condition and determine what to discuss. Two datasets, MedDG \cite{meddg} and KaMed \cite{vrbot}, are annotated following the few-shot prompt in \S \ref{align}. As shown in Table \ref{thought_corpus}, the former is annotated with 122K thought processes and the latter with 285K. Each thought process includes a reasoning path consisting of approximately four steps. An example of the generated thought process is displayed in Table \ref{thought_example}. The doctor investigates the new clinical findings (i.e., discomfort under certain conditions) and plans to discuss allergy issues after multi-step reasoning. 

Human assessments are carried out to evaluate the quality of the generated thought processes. We randomly select 100 samples from each dataset and ask three medical students who have undergone clinical internships to assess them. The evaluation employs three metrics: (1) \textbf{Knowledge}: whether the knowledge used in the thought process is accurate; (2) \textbf{Consistency}: whether the thought process is consistent with the dialogue history; (3) \textbf{Rationality}: whether the thought process starts with premises and uses logical progression to derive responses. Table \ref{human_cot} presents the evaluation results, demonstrating that the thought processes in MEDIATOR adequately fulfill the above three criteria. 

\section{Experiments}

\begin{table*}[t!]
\center
\resizebox{0.8\textwidth}{!}{
\begin{tabular}{@{}clcccccccccc@{}}
\toprule
\multicolumn{2}{l}{}                              & \multicolumn{5}{c}{\textbf{MedDG}}             & \multicolumn{5}{c}{\textbf{KaMed}}             \\ \cmidrule(l){3-12} 
\multicolumn{2}{c}{\textbf{Methods}}                       & B-1   & B-4   & R-1   & R-2   & E-F   & B-1   & B-4   & R-1   & R-2   & E-F   \\ \midrule
\multicolumn{1}{c}{\multirow{2}{*}{Zero-shot LLMs}}  & HuatuoGPT-II & 42.45 & \textbf{24.78} & 15.85 & 4.24  & 9.45  & 40.89 & 22.9  & 18.06 & 4.65  & 11.46 \\
                                   & DISC-MedLLM  & 40.72 & 22.6  & 20.13 & 6.6   & 10.15 & 38.05 & 20.26 & 20.48 & 5.93  & 13.54 \\ 
                                   & GPT-4  & 42.19 & 23.32  & 13.99	 & 3.47   & 13.15 & \textbf{41.88}  & \textbf{23.34}  & 13.94  & 3.1    &  13.86 \\ \midrule
\multicolumn{1}{c}{\multirow{6}{*}{Fine-tuned Models}} & VRBot        & 29.69 & 16.34 & 24.69 & 11.23 & 12.78 & 30.04 & 16.36 & 18.71 & 7.28  & 12.08 \\
                                   & GPT-2        & 35.27 & 19.16 & 28.74 & 13.61 & 16.14 & 33.76 & 17.82 & 26.80 & 10.56 & 17.26 \\
                                   & BART         & 34.94 & 19.06 & 29.03 & 14.40 & 16.66 & 33.62 & 17.64 & 27.91 & 11.43 & 19.20 \\
                                   & Qwen-7B      & 35.11 & 19.03 & 30.19 & 15.01 & 18.05 & 34.00 & 17.66 & 28.34 & 12.18 & 19.88 \\
                                   & DFMed        & \textbf{42.83} & 22.90 & 29.72 & 14.31 & 22.92 & 40.50 & 20.92 & 28.33 & 11.73 & 22.31 \\
                                   & \textsc{Emulation}    & 42.35 & 22.76 & \textbf{30.91}$^\dag$ & \textbf{15.17}$^\dag$ & \textbf{24.03}$^\dag$ & 39.87 & 19.79 & \textbf{28.54}$^\dag$ & \textbf{12.33}$^\dag$ & \textbf{24.27}$^\dag$ \\ \bottomrule
\end{tabular}
}
\caption{Automatic evaluation results on two datasets. † denotes statistically significant differences ($p < 0.05$).}
\label{main_result}
\end{table*}

\subsection{Datasets}

Our experiments utilize two medical dialogue datasets, MedDG \cite{meddg} and KaMed \cite{vrbot}. Dialogues in these datasets exhibit a clear multi-turn context, with each dialogue averaging about 10 turns. The MedDG dataset comprises 17,860 dialogues, focusing on 12 gastroenterology-related diseases. The dataset is divided into 14,862/1,999/999 for training, validation, and testing. The KaMed dataset includes more than 63,000 dialogues, spanning a wide range of diseases across approximately 100 hospital departments. We clean privacy-sensitive content following DFMed \cite{dfmed} and divide the dataset into 29,159/1,532/1,539 for training, validation, and testing.

\subsection{Baseline methods}

We compare our method with two categories of baselines: LLMs equipped with Chinese medical conversation abilities and language models that are fine-tuned on target datasets. 
\paragraph{Medical LLMs.}
(1) \textbf{HuatuoGPT-II} \cite{huatuo2} is a medical LLM performing state-of-the-art in several Chinese medical tasks. (2) \textbf{DISC-MedLLM} \cite{disc_medllm} is a medical LLM with strong multi-turn consultation capabilities. (3) \textbf{GPT-4} \cite{gpt-4} is one of the most advanced pre-trained LLMs designed by OpenAI. 

\paragraph{Fine-tuned Models.} 
(1) \textbf{VRBot} \cite{vrbot} is a medical dialogue generation (MDG) model with entity tracking and predicting. (2) \textbf{GPT-2} \cite{gpt-2} is a transformer decoder-based language model. (3) \textbf{BART} \cite{bart} is a transformer-based encoder-decoder model. (4) \textbf{Qwen-7B} \cite{qwen} is a strong base language model focusing on Chinese and English. (5) \textbf{DFMed} \cite{dfmed} enhances MDG with entity and dialogue act flow learning.

\subsection{Implementation Details}

All the prompt-based operations in \textsc{Emulation} are implemented with \texttt{gpt-3.5-turbo}, i.e., clinical findings extractions, abductive diagnosis refinement, and deductive diagnosis analysis. 
We apply the MedBERT\footnote{https://github.com/trueto/medbert} pre-trained in the medical domain as the backbone of the disease retriever and the disease alignment model. The corpus of disease documents, utilized for disease retrieval and to enhance both abductive and deductive reasoning, is derived from a specialist-certified online medical knowledge base \textit{xiaohe}\footnote{https://www.xiaohe.cn/medical}. We retrieve the top 50 ($K$=50) relevant diseases for further refinement. The refined disease list is sent to the disease alignment model to obtain the top 5 ($K''$=5) diseases that may be discussed in subsequent responses. 
Then, we employ the pre-trained language model Qwen-7B-Chat\footnote{https://github.com/QwenLM/Qwen} to train the thought process alignment model, which has seven billion parameters and proficient Chinese understanding and generation ability. We train the model using the LoRA approach with $r$=64 and $\alpha$=16. All experiments are carried out on a system equipped with four RTX 3090 GPUs. Other details are presented in the Appendix \ref{implement}.

\subsection{Automatic Evaluation}

\begin{table}[]
\resizebox{1\linewidth}{!}{
\begin{tabular}{@{}lccccc@{}}
\toprule
\textbf{Methods}     & \textbf{Proactivity} & \textbf{Accuracy} & \textbf{Helpfulness} & \textbf{LQ} & \textbf{Average} \\ \midrule
DISC-Med  & 4.52        & 4.41     & 4.71        & 4.98               & 4.66    \\
DFMed     & 4.73        & 4.48     & 4.79        & 4.86               & 4.72    \\
EMULATION & 4.69        & 4.65     & 4.82        & 4.99               & 4.79    \\ \bottomrule
\end{tabular}
}
\caption{Automatic evaluation results based on GPT-4.}
\label{gpt-4-eval}
\end{table}

We assess the generated responses using three automated metrics: BLEU-1/2/4 (\textbf{B-1/4}) \cite{bleu}, evaluating $n$-gram precision; ROUGE-1/2 (\textbf{R-1/2}) \cite{rouge}, assessing $n$-gram recall; and Entity-F1\footnote{https://github.com/lwgkzl/MedDG} (\textbf{E-F}) \cite{meddg}, which gauges the accuracy of medical entities, such as diseases and medications. 

Table \ref{main_result} displays the response generation results of all baseline methods. Our framework outperforms the baselines in most metrics, especially the medical entity accuracy. It demonstrates the effectiveness of our framework in diagnostic reasoning and response generation. Specifically, \textsc{Emulation} performs better than available medical LLMs trained on large-scale medical dialogues and knowledge bases in R-1/2 and E-F. It demonstrates that modeling the diagnostic reasoning process can help generate more accurate and targeted responses. Besides, our framework achieves better $n$-gram recall and entity accuracy than the state-of-the-art model DFMed. The B-1/4 scores are a bit lower, which may be because the fine-tuned parameters are too limited to approach the linguistic pattern of the gold utterances. However, benefiting from the construction of diagnostic reasoning processes, \textsc{Emulation} can generate responses with more consistent content and higher entity accuracy. 

We also adopt another multi-dimensional automatic evaluation using GPT-4, which follows the method in DISC-Med \cite{disc_medllm}. This evaluation focuses on \textbf{Proactivity}, \textbf{Accuracy}, \textbf{Helpfulness}, and \textbf{Linguistic Quality (LQ)}. Our \textsc{Emulation} performs better than other baselines in the above aspects as shown in Table \ref{gpt-4-eval}. 

\subsection{Human Evaluation}

\begin{figure}[t!]
	\centering
	\includegraphics[width=1\linewidth]{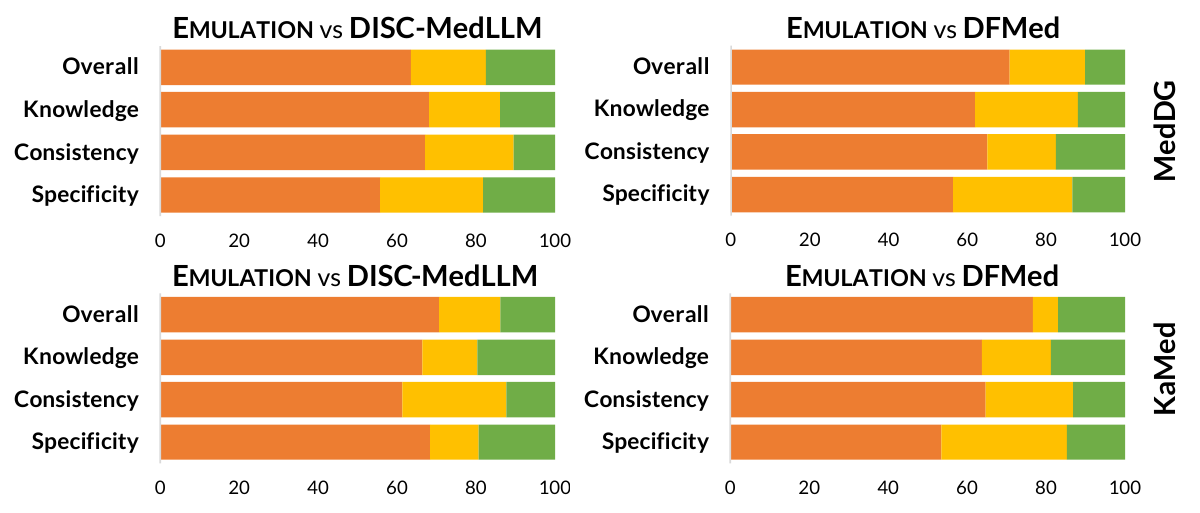}
	\caption{Human evaluation results of baseline methods. Red is for \textbf{Win}, yellow for \textbf{Tie}, and green for \textbf{Lose}.}
	\label{human}
\end{figure}

We conduct human evaluations on comparison with two baselines: DISC-MedLLM and DFMed. We randomly select 100 samples and ask three medical students to evaluate them according to \textbf{Knowledge}, \textbf{Consistency}, and \textbf{Specificity}. Besides, the overall quality is compared. \textsc{Emulation} outperforms two baselines in all aspects, as shown in Figure \ref{human}. Details and case studies are in the Appendix \ref{human_eval_baseline} and \ref{case_study}. Cross-inference cases are displayed.

\subsection{Analysis of Diagnostic Reasoning Process}

\begin{table}[]
\center
\resizebox{1\linewidth}{!}{
\begin{tabular}{@{}lllllll@{}}
\toprule
                        & \multicolumn{3}{c}{\textbf{MedDG}} & \multicolumn{3}{c}{\textbf{KaMed}} \\ \cmidrule(l){2-7} 
\textbf{Methods}                 & B-4     & R-2    & E-F    & B-4     & R-2    & E-F    \\ \midrule
\textsc{Emulation}      & \textbf{22.76}   & \textbf{15.17}  & \textbf{24.03}  & \textbf{19.79}   & \textbf{12.33}  & \textbf{24.27}  \\
w/o Abd. Reasoning      & 20.30   & 13.77  & 17.47  & 18.02   & 11.31  & 19.75  \\
w/o Ded. Reasoning      & 22.34   & 15.07  & 23.77  & 19.63   & 12.05  & 24.11  \\
w/o Dis. Alignment      & 21.93   & 14.69  & 21.24  & 18.84   & 11.87  & 22.43  \\ 
w/o Thot. Alignment     & 22.31   & 14.95  & 23.82  & 19.72   & 12.12  & 24.08  \\ \bottomrule
\end{tabular}
}
\caption{Ablation study on two datasets}
\label{ablation}
\end{table}

To further evaluate the effectiveness of our framework, we analyze several variations of our \textsc{Emulation} as detailed below:
(1) \textbf{w/o Abd. Reasoning}, which omits abductive reasoning along with any related deductive reasoning, relying solely on dialogue history to form thought processes.
(2) \textbf{w/o Ded. Reasoning}, which eliminates the process of deductively analyzing clinical findings and potential diseases.
(3) \textbf{w/o Dis. Alignment}, which forgoes learning the disease priority, opting to select the most relevant diseases identified through abductive reasoning.
(4) \textbf{w/o Thot. Alignment}, which directly generates responses based on the results of abductive and deductive reasoning. 

Table \ref{ablation} presents the comprehensive results of our ablation study. There is a noticeable decline in effectiveness across various metrics for the ablation models, underscoring the indispensable contribution of each module within our framework. Particularly, the variant \textit{w/o Abd. Reasoning} experiences a significant drop in response quality due to the lack of initial diagnosis and subsequent analysis. This decline is attributed to the typical practice of clinicians to center medical conversations around certain diseases; thus, omitting precise abductive and deductive reasoning processes leads to a lack of focus in the dialogue. Besides, the \textit{w/o Dis. Alignment} variant shows a marked decrease in performance, reinforcing the importance of disease priority alignment with clinician practices for achieving consistent responses. The outcomes from \textit{w/o Ded. Reasoning} and \textit{w/o Thot. Alignment} further affirm that detailed analyses and thought generation enhance the quality of response generation.

\subsection{Analysis of Disease Alignment}

\begin{figure}[t!]
	\centering
	\includegraphics[width=0.95\linewidth]{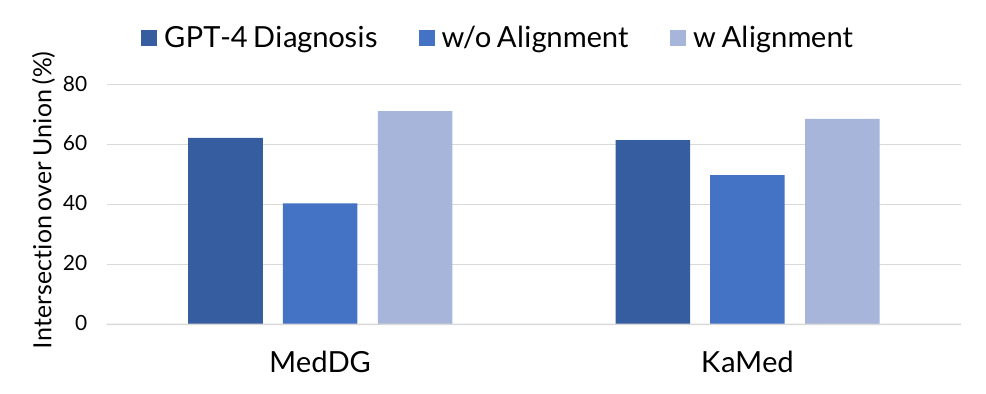}
	\caption{The top-5 diagnosis results with or without disease priority alignment.}
	\label{iou}
\end{figure}

\begin{figure}[t!]
	\centering
	\includegraphics[width=0.85\linewidth]{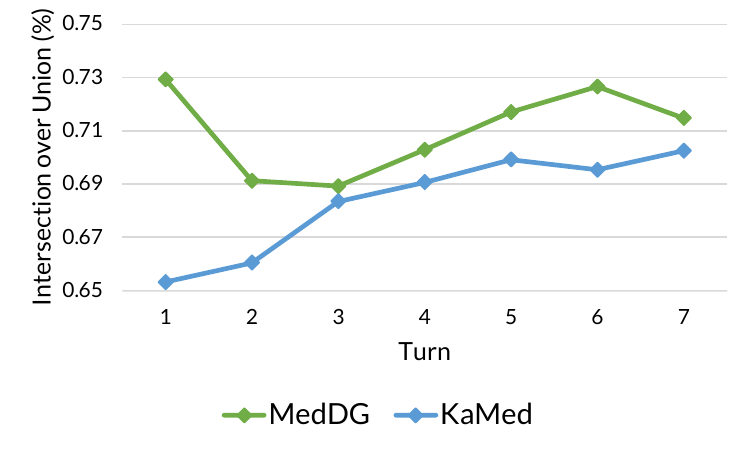}
	\caption{The top-5 diagnosis results of \textsc{Emulation} at different conversation turns.}
	\label{turn_iou}
\end{figure}

We evaluate the diagnosis accuracy to determine the impact of aligning disease priorities. The metric Intersection over Union (IoU) between predicted diseases and plausible diseases $E_t^{post}$ is employed. As shown in Figure \ref{iou}, aligned diagnoses substantially outperform unaligned diagnoses that are derived directly from abductive reasoning. Remarkably, aligned diagnoses are marginally more accurate than the diseases diagnosed by GPT-4, i.e., $E_t^{pri}$. We also display the diagnosis results in different conversation turns as shown in Figure \ref{turn_iou}. Overall, the diagnosis accuracy continuously improves as the number of conversation turns increases. The diagnostic accuracy on the MedDG dataset was initially high but decreased in the second turn. It is because the patients in this dataset provide relatively sufficient information in the first turn, allowing \textsc{Emulation} to make a good rough diagnosis. However, with the introduction of additional information, the difficulty of diagnosis increases, and the difference between \textsc{Emulation}’s diagnosis and that of real doctors becomes greater.

\section{Related Work}

\paragraph{Medical Dialogue Systems}
Medical Dialogue Systems (MDS) aim to provide healthcare services to patients. A significant and early area of research focuses on automated diagnosis in the form of task-oriented dialogue systems, which prioritize the quick identification of underlying symptoms and the provision of a final diagnosis and do not offer further consultations \cite{tod-diagnosis-hrl,symp-graph,diaformer,br_agent}. The study by \citet{tod-diagnosis} presented a dataset with symptom annotations and established an MDS using reinforcement learning. \citet{kg-routed-diagnosis} incorporated a knowledge graph into MDS to control the sequence of symptom inquiries. \citet{trust_med} enhanced system dependability by applying an exploration-confirmation approach and giving precedence to severe diseases. 

The emergence of large-scale medical dialogue datasets such as MedDialog \cite{meddialog}, MedDG \cite{meddg}, and KaMed \cite{vrbot}, along with pre-trained language models \cite{bart,gpt-2}, has sparked increased interest in medical dialogue generation \cite{meddg,hetero,geml,med_pivotal}. The research by \citet{meddg} tackled medical dialogue generation by emphasizing entity prediction and entity-centric response creation. \citet{vrbot} presented a semi-supervised variation reasoning system supplemented by a patient state tracker and a physician action network. \citet{dfmed} introduced a dual flow (i.e., dialogue act and entity flows) modeling approach to enhance dialogue understanding and guide response generation using acts and entities. \citet{plugmed} applies LLMs to medical dialogue generation in a plug-and-play way. 

\paragraph{Medical Large Language Models}
Given the astonishing performance of GPT-4 in several medical examinations, an increasing number of researchers are directing their attention toward developing medical LLMs. ChatDoctor \cite{chatdoctor} is equipped with an external Wikipedia knowledge base and trained on real medical conversations. DoctorGLM \cite{doctorglm} is developed using a medical dialogue and question answering dataset supplemented by ChatGPT-translated documents. HuatuoGPT-2 \cite{huatuo2} and DISC-MedLLM \cite{disc_medllm} try to construct a unified domain adaption framework that uses ChatGPT to convert available documents into pre-training and fine-tuning instructions. 

Available medical dialogue systems and medical LLMs try to learn from the outcome of the diagnostic reasoning (i.e., high-quality medical dialogue datasets) but ignore the internal thought process of real clinicians and alignment with clinician preferences. Our work seeks to construct a medical dialogue system that aligns with the internal diagnostic reasoning process of real clinicians.

\section{Conclusion}

This paper proposes a novel medical dialogue system framework, \textsc{Emulation}, that emulates clinicians' diagnostic reasoning processes to generate appropriate responses grounded in abductive and deductive diagnostic reasoning analysis and alignment with clinician preferences. Besides, a new diagnostic thought process corpus is presented and utilized to model the clinician preference. Experimental results demonstrate the effectiveness of \textsc{Emulation} on two datasets. One promising area for future work is applying the \textsc{Emulation} framework in telemedicine consultations. With the increasing demand for telemedicine, it is essential to explore how this framework can enhance virtual patient-doctor interactions. Future studies could investigate the framework’s effectiveness in improving patient satisfaction and overall consultation quality in remote settings.

\section*{Limitations}

While our framework outperforms various baseline approaches in medical dialogue generation, there is still room for progress. The corpus for the diagnostic thought process is constructed by inferring from doctors' responses. Although it has been assessed that the thought processes in the corpus demonstrate a logical sequence, individual steps within these processes might not mirror those of a real clinician precisely. To enhance consistency in generated thought processes, additional human-annotated thought processes should be collected to conduct further alignment.

\section*{Ethics Statement}

Our designed system is intended to improve medical consultations for patient care. All datasets were anonymized upon their publication in dataset papers. Nonetheless, due to the training of our model on a limited number of samples for certain conditions, there's a possibility that the responses might contain inaccurate information regarding diagnosis, treatment, and safety measures. We advise treating our system as a supplementary resource and seeking professional medical advice when necessary. Additionally, user interactions with the system could potentially expose sensitive data (e.g., user-reported gender), and the online LLM API services should be substituted by a local open-source model if our system is deployed. Therefore, we urge users to meticulously assess the ethical considerations of the generated responses. Moreover, the scientific tools utilized in our research, such as NLTK, ROUGE, Transformers, and various GitHub repositories, are openly accessible for academic purposes. The application of these tools in this study adheres to their designated purposes.


\section*{Acknowledgment}
This work was supported by the Research Grants Council of Hong Kong (15207920, 15213323) and the National Natural Science Foundation of China (62076212). 

\bibliography{custom}

\appendix

\section{Appendix}

\subsection{Automatic Evaluation Details}

We adopt the calculation approach used in the original dataset paper MedDG \cite{meddg} and in the most recent paper DFMed \cite{dfmed}. The “nltk” package with version 4.5.1 is used to calculate BLEU scores. The “rouge” version is 1.0.1. There is a score gap between our study and some baseline studies since the metric package used for calculating BLEU is different. For example, the officially released code for VRBot doesn’t specify the metric package version such as “nlgeval” and “rouge”, while the results will be affected by the package version. 

\subsection{Disease Annotation}\label{disease_annotate}

\begin{figure*}[!t]
	\centering
	\includegraphics[width=1\linewidth]{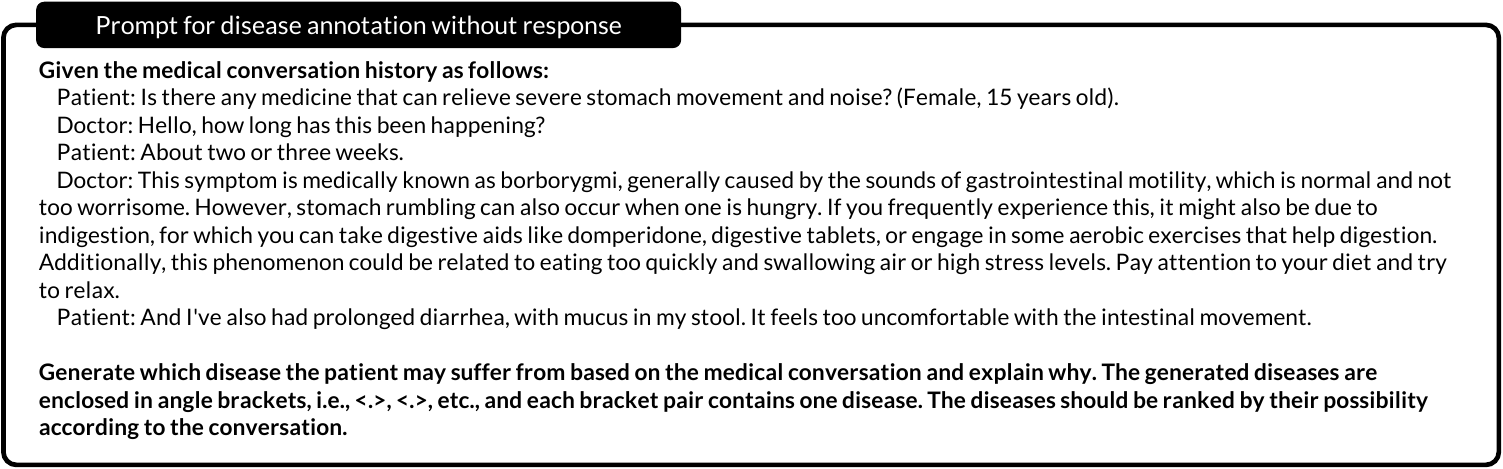}
	\caption{Prompt for disease annotation without the doctor's response.}
	\label{prompt_disease_annotate_wo}
\end{figure*}

\begin{figure*}[!t]
	\centering
	\includegraphics[width=1\linewidth]{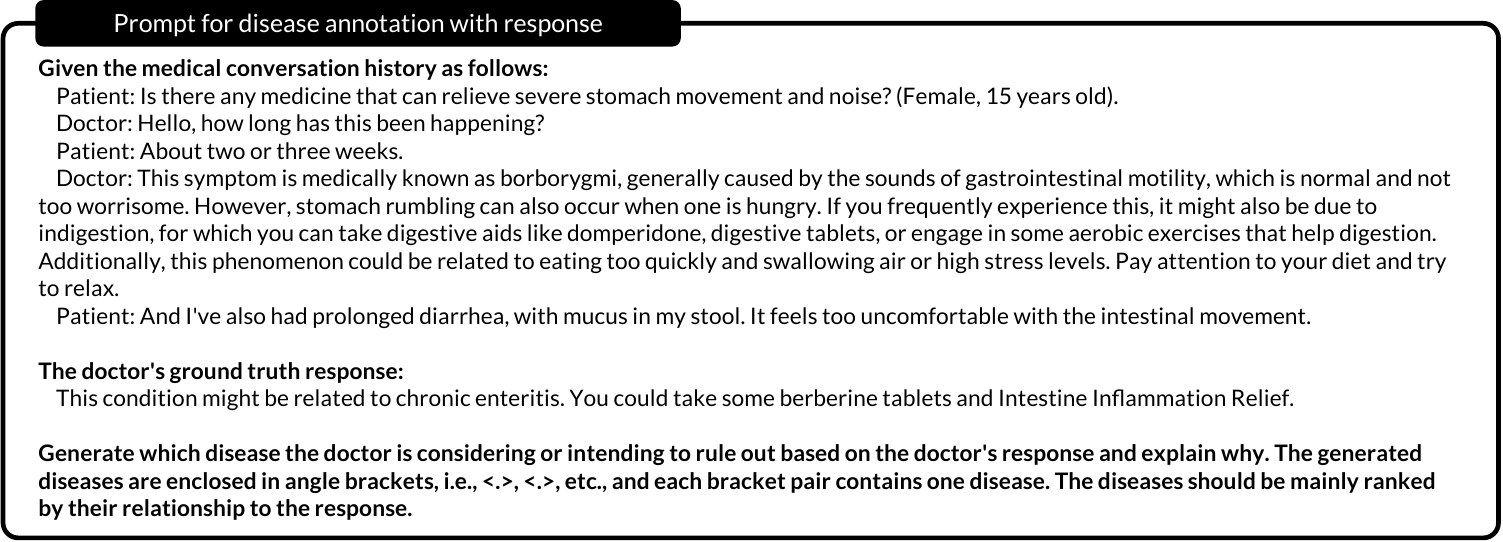}
	\caption{Prompt for disease annotation with the doctor's response.}
	\label{prompt_disease_annotate_w}
\end{figure*}

We annotate the potential diseases for each dialogue turn with the help of GPT-4 using two different prompts. One is ``Generate which disease the patient may suffer from based on the medical conversation and explain why. The diseases should be ranked by their possibility according to the conversation.'' This prompt aims to leverage the diagnostic capability of GPT-4 to infer potential diseases. The other is ``Generate which disease the doctor is considering or intending to rule out based on the doctor's response and explain why. The diseases should be mainly ranked by their relationship to the response.'' This prompt aims to deduce diseases from the doctor's response. The difference between these two prompts is whether the current doctor's response is given. Examples of two disease annotation prompts are shown in Figure \ref{prompt_disease_annotate_wo} and Figure \ref{prompt_disease_annotate_w}.

\begin{figure*}[!t]
	\centering
	\includegraphics[width=1\linewidth]{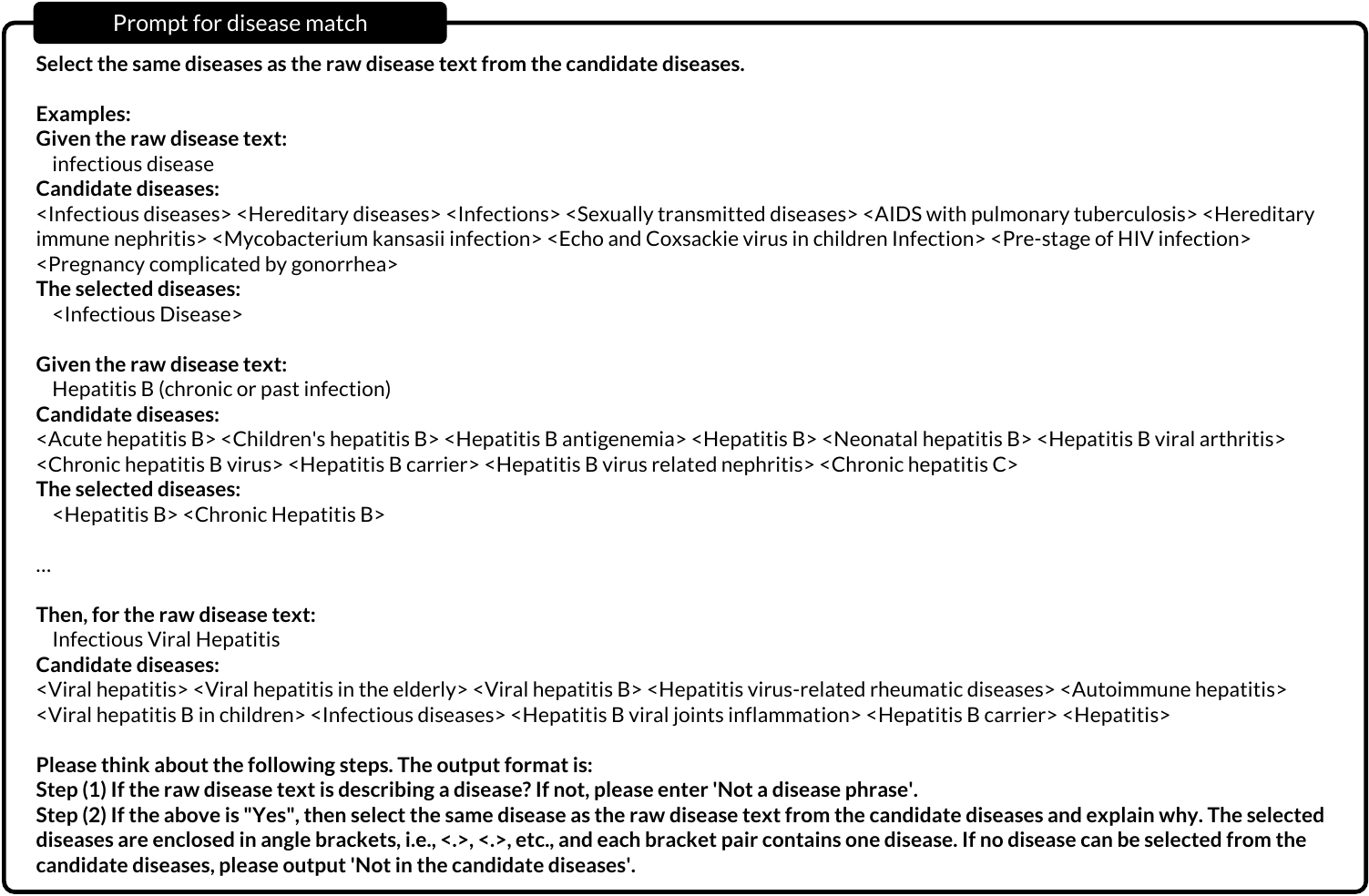}
	\caption{Prompt for disease match.}
	\label{prompt_disease_match}
\end{figure*}

After inferring diseases using two prompts respectively, we need to link each inferred disease to the disease in an external knowledge base. We divide this linking into two steps: (1) Coarse matching and (2) GPT-4 assisted matching. For the coarse matching, we first build a dense retriever to calculate the relevance score between disease name and disease documents. The disease documents are from the knowledge base. This retriever can bridge the gap between different aliases of one disease. Then, we retrieve the top 10 relevant diseases as a preliminary list. For the GPT-4 assisted matching, we organize a prompt that requires GPT-4 to select from the preliminary disease list. The prompt is ``Select diseases from the candidate disease list that describe the same one as our target.'' An example of a disease match prompt is shown in Figure \ref{prompt_disease_match}.

\subsection{Implementation Details}\label{implement}

\paragraph{Baseline Methods.}
For the medical LLMs, we use the HuatuoGPT2-13B\footnote{https://github.com/FreedomIntelligence/HuatuoGPT-II} and DISC-MedLLM\footnote{https://github.com/FudanDISC/DISC-MedLLM} to generate responses in zero-shot way as they have already been fine-tuned on medical dialogue datasets and other large-scale medical knowledge datasets. These two models are all based on language models with 13B parameters. We utilize the default decoding parameters to generate responses. 

For fine-tuned baseline methods, we employ generation results from the original papers. 

\paragraph{Fine-tuned modules in our framework.}
For training disease retriever in \S \ref{abductive}, we use the MedBERT as the encoder and employ contrastive learning. The batch size is 8 with 6 gradient accumulation steps. The learning rate is set at 3e-5. We train 6 epochs for the MedDG dataset and 10 epochs for the KaMed dataset. We select the checkpoint with the highest disease recall rate in validation datasets. 

For training the disease priority alignment model, we also use the MedBERT as the encoder and employ contrastive learning. The batch size is 2 with 8 gradient accumulation steps. The learning rate is set at 1e-5. We train 3 epochs for two datasets and select the checkpoint with the highest IoU in validation datasets.

For training the thought process generation model, we use the Qwen-7B-Chat model and adopt LoRa fine-tuning. The LoRa settings are: $r$ equals 64, $\alpha$ equals 16, dropout rate equals 0.05. The learning rate is set at 3e-4, the batch size for each GPU is 2 with 8 gradient accumulation steps. We train the LoRa parameters for 3 epochs and select the last checkpoint to generate responses. 

\begin{figure*}[!t]
	\centering
	\includegraphics[width=1\linewidth]{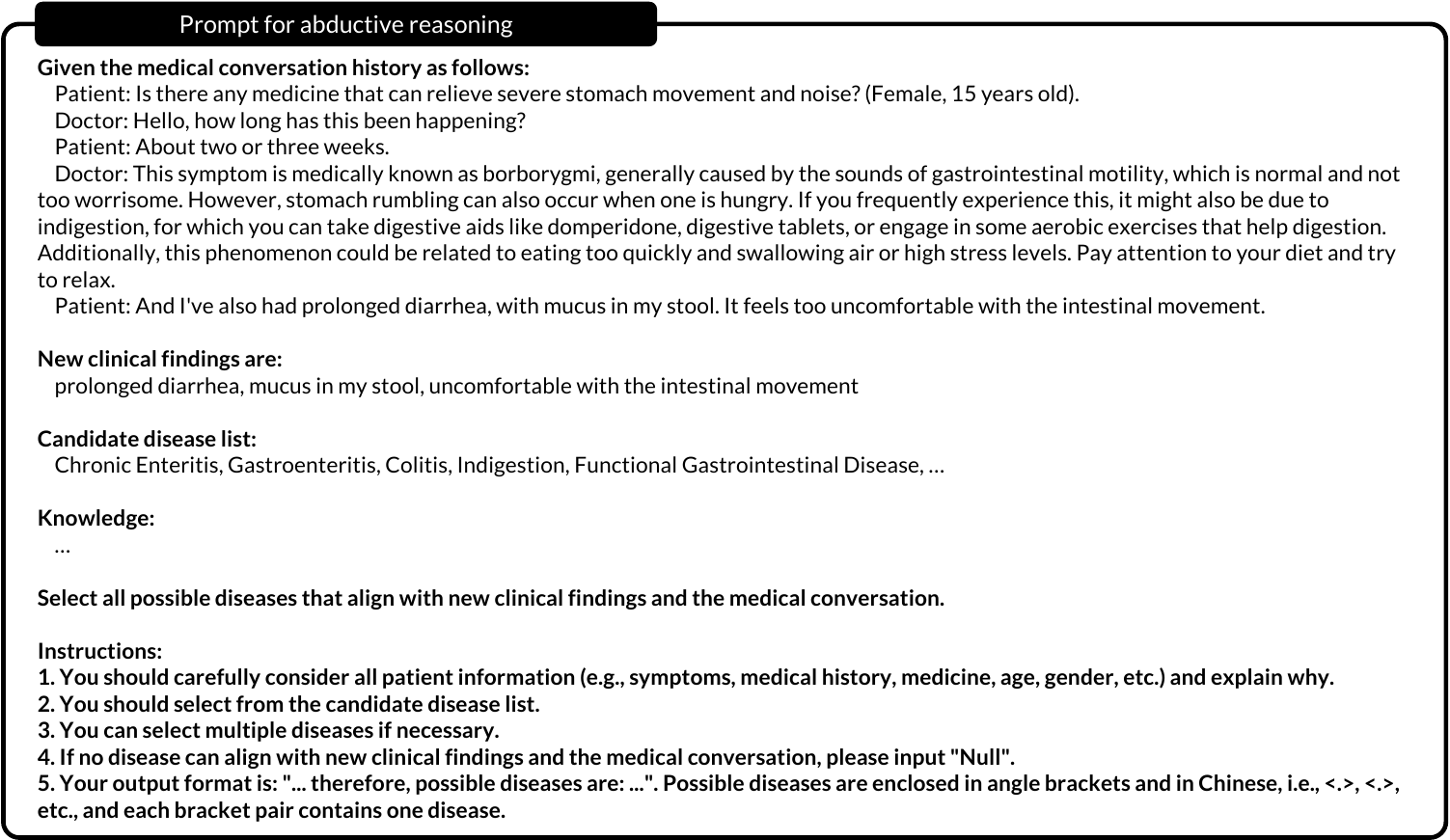}
	\caption{Prompt for abductive reasoning.}
	\label{prompt_abductive}
\end{figure*}

\begin{figure*}[!t]
	\centering
	\includegraphics[width=1\linewidth]{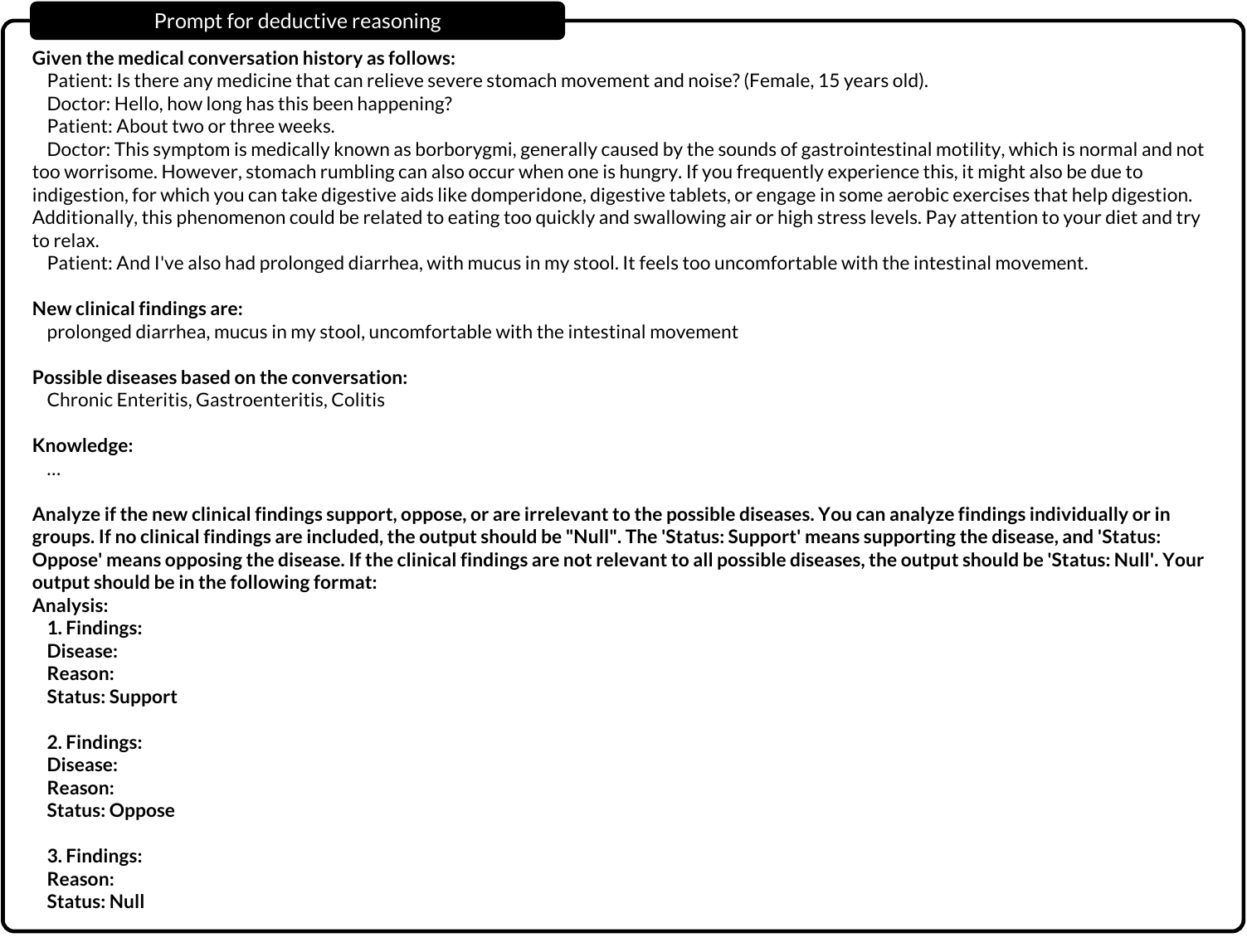}
	\caption{Prompt for deductive reasoning.}
	\label{prompt_deductive}
\end{figure*}

\paragraph{Prompt modules in our framework.}
For the abductive reasoner, the prompt is shown in Figure \ref{prompt_abductive}. For the deductive reasoner,  the prompt is shown in Figure \ref{prompt_deductive}.

\subsection{Disease Retrieval Results}

\begin{table}[t!]
\center
\resizebox{0.9\linewidth}{!}{
\begin{tabular}{ccccc}
\toprule
\textbf{Datasets}  & \textbf{Top-10}  & \textbf{Top-25}  & \textbf{Top-50} & \textbf{Top-100} \\ \midrule
MedDG    & 60.90\%        & 83.99\%         & 92.88\%     & 96.85\%        \\
KaMed    & 64.91\%        & 83.10\%         & 90.64\%  & 95.04\%       \\ \bottomrule
\end{tabular}
}
\caption{Retrieval results.}
\label{retrieval_results}
\end{table}

Table \ref{retrieval_results} displays the recall rate of different top-$K$ diseases (i.e., $K$=10, 25, 50, 100). 

\subsection{Human Evaluations on Comparison with Baseline Methods}\label{human_eval_baseline}

We compare our framework \textsc{Emulation} with baseline methods from three aspects. (1) \textbf{Knowledge Accuracy (Knowledge)}, which assesses whether the response correctly applies the disease knowledge. (2) \textbf{Consistency}, which assesses whether the response is consistent with the dialogue context and doctor's ground truth response. (3) \textbf{Specificity}, which assesses whether the response provides specific diagnoses, prescriptions, treatment plans, or examination requirements rather than general suggestions. We ask our annotators to compare the responses generated by our method with the responses generated by baseline methods.

\subsection{Case Studies}\label{case_study}

\begin{figure*}[!t]
	\centering
	\includegraphics[width=1\linewidth]{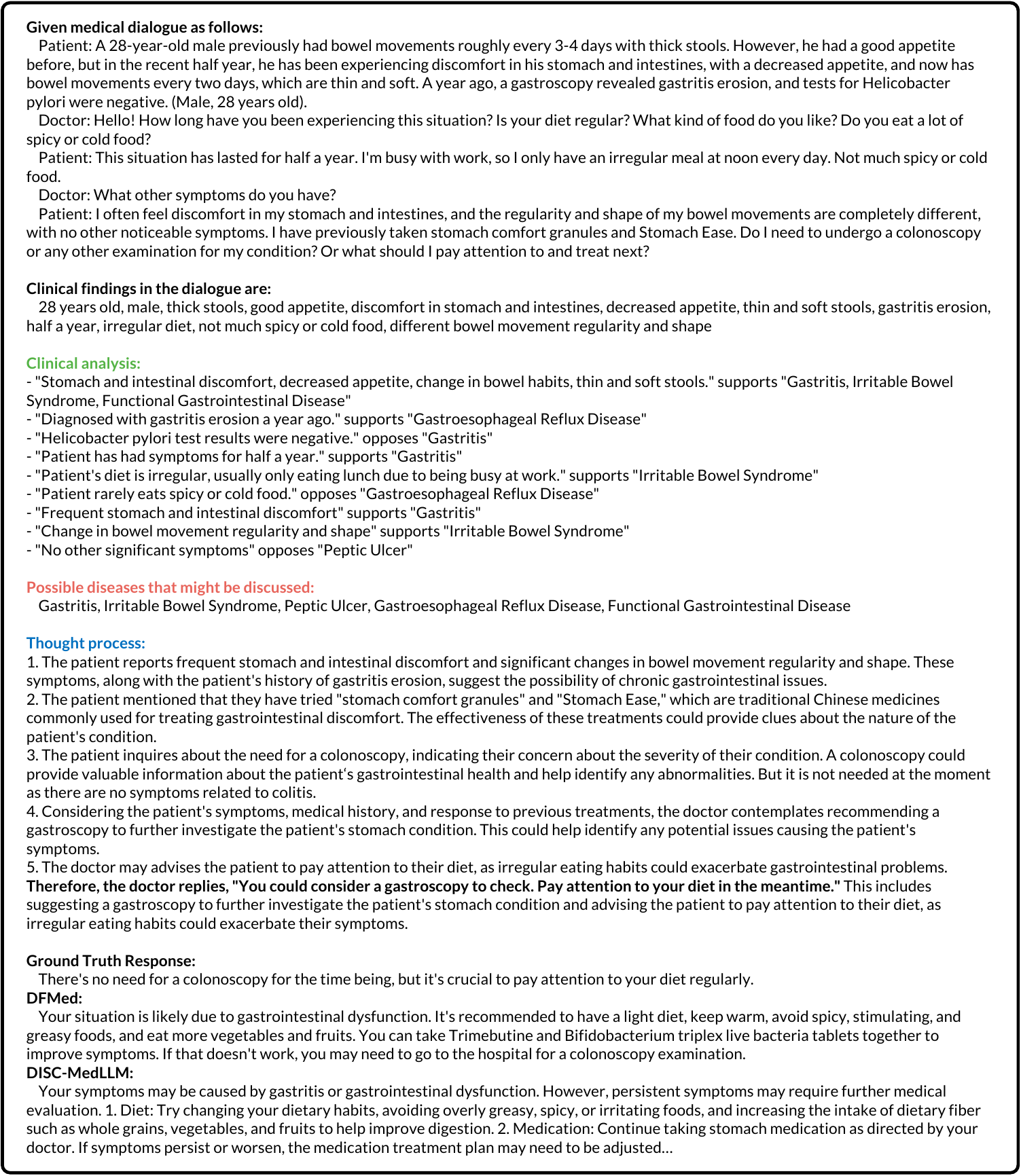}
	\caption{Generated case on the MedDG dataset.}
	\label{case_meddg}
\end{figure*}

\begin{figure*}[!t]
	\centering
	\includegraphics[width=1\linewidth]{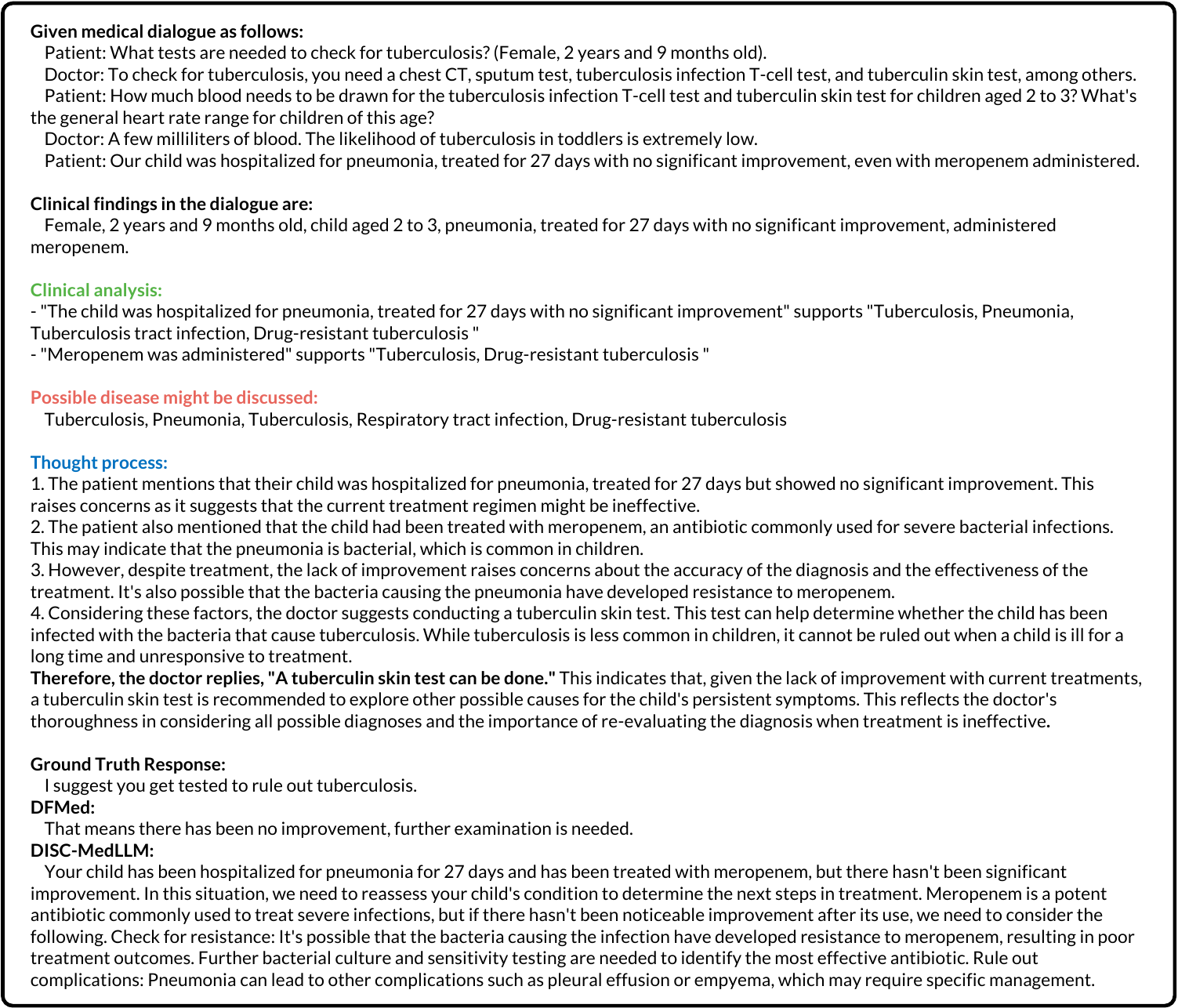}
	\caption{Generated case on the KaMed dataset.}
	\label{case_kamed}
\end{figure*}

\begin{figure*}[!t]
	\centering
	\includegraphics[width=1\linewidth]{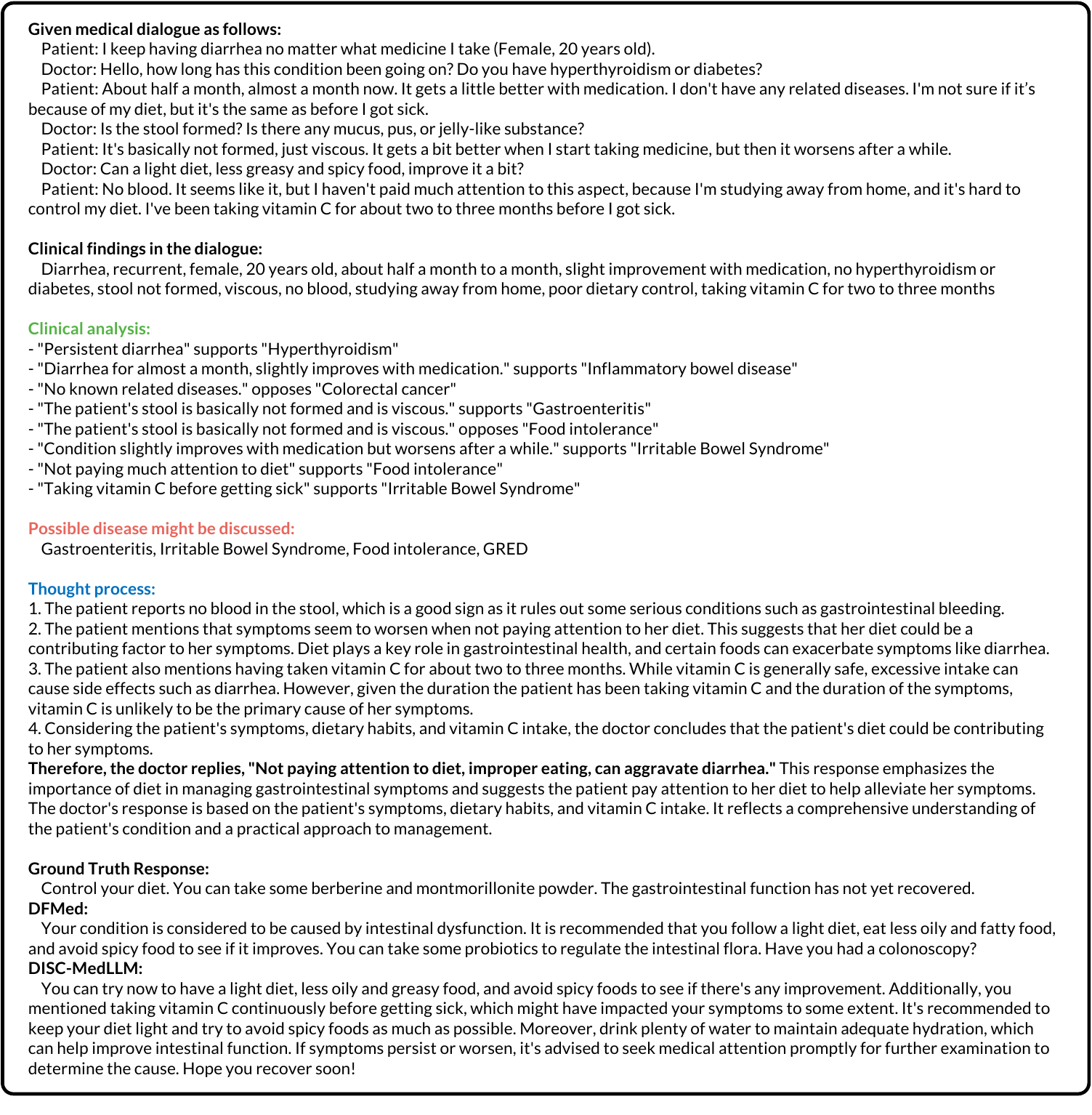}
	\caption{Generated MedDG case with the model trained on KaMed.}
	\label{case_meddg_cross}
\end{figure*}

\begin{figure*}[!t]
	\centering
	\includegraphics[width=1\linewidth]{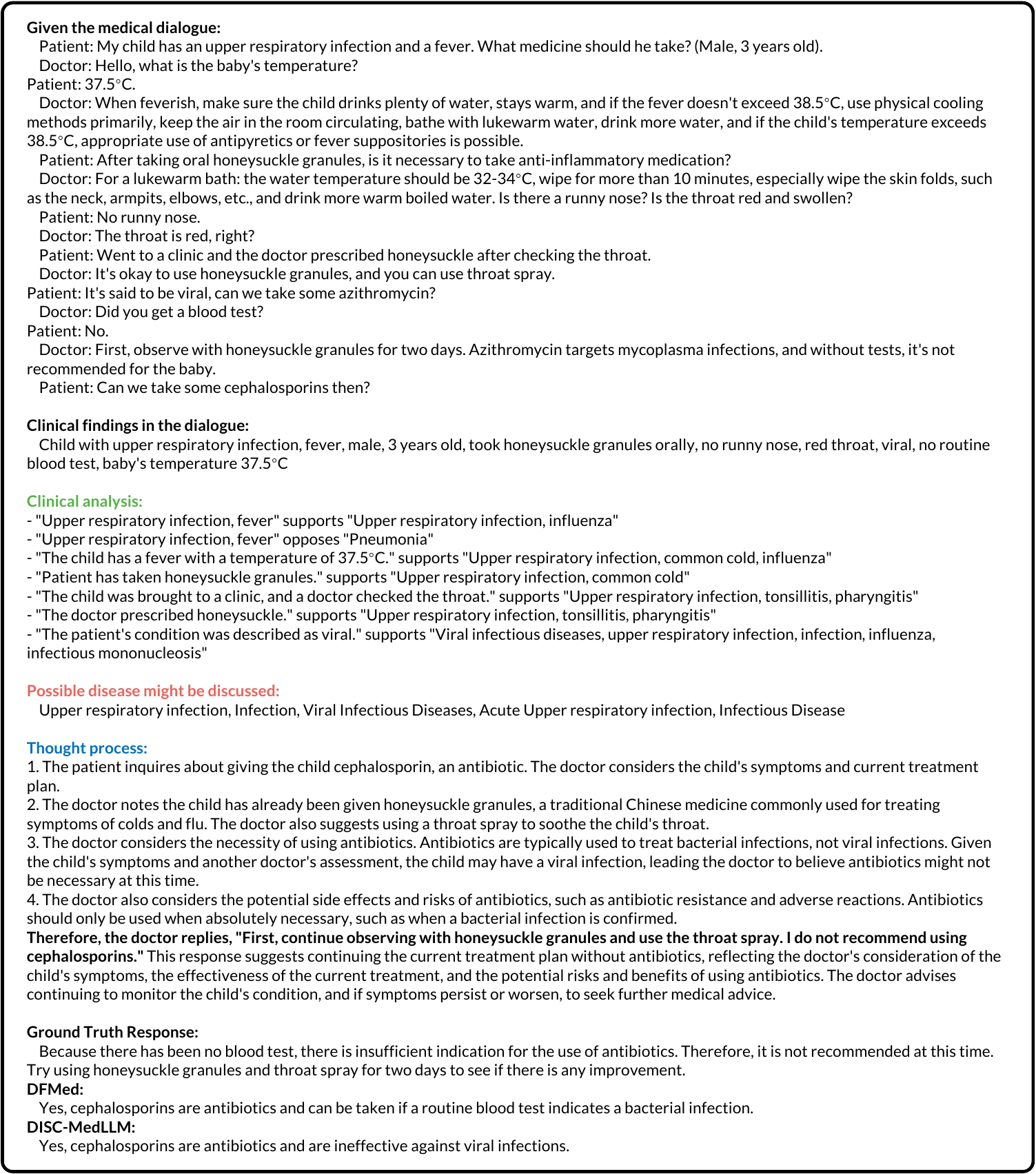}
	\caption{Generated KaMed case with the model trained on MedDG.}
	\label{case_kamed_cross}
\end{figure*}

Figure \ref{case_meddg} and Figure \ref{case_kamed} display cases on the MedDG and KaMed datasets, respectively. Besides, we also infer samples in MedDG with the model trained on KaMed and infer samples in KaMed with the model trained on MedDG in Figure \ref{case_meddg_cross} and Figure \ref{case_kamed_cross}.

\end{document}